\documentclass[11pt]{article}

\usepackage[final]{acl}

\usepackage{times}
\usepackage{latexsym}

\usepackage[T1]{fontenc}

\usepackage[utf8]{inputenc}

\usepackage{microtype}

\usepackage{inconsolata}

\usepackage{graphicx}
\usepackage{amssymb}  

\usepackage{kotex}
\usepackage{booktabs}
\usepackage{amsmath}

\title{Reducing Logprob Without Causal Verification: Diagnosing SAE-Based PII Suppression in LLMs}

\author{Chaerin Lee\textsuperscript{*}, Wonho Lee\textsuperscript{*}, Sohee Park, Daeseon Choi \\
        Soongsil University \\
        {\{chaerin1112, hoho0907, sosohi}\}@soongsil.ac.kr, sunchoi@ssu.ac.kr
        }

\hypersetup{
  pdftitle={Reducing Logprob Without Causal Verification: Diagnosing SAE-Based PII Suppression in LLMs},
  pdfauthor={Chaerin Lee, Wonho Lee, Sohee Park, Daeseon Choi}
}

\begin{document}
\maketitle
\renewcommand{\thefootnote}{}\footnotetext{\textsuperscript{*}~~These authors contributed equally and are listed in alphabetical order.\\
Corresponding Author: Daeseon Choi, sunchoi@ssu.ac.kr}
\renewcommand{\thefootnote}{\arabic{footnote}}
\begin{abstract}
    SAE-based feature-level interventions to suppress PII leakage in LLMs are primarily evaluated by log-probability reduction. We show that this criterion cannot distinguish feature-specific suppression from a magnitude artifact: a direction-agnostic effect of sufficiently large activation perturbations. Across 36 (model, layer, PII type) cases, we compare selected SAE feature interventions with coherence-matched random direction controls using logprob and rank-aware metrics. While 17 cases (47\%) appear successful under logprob-only evaluation, only 1 case (2.8\%) passes our multi-metric framework. The rest reveal two failure modes: random directions outperform selected features, and feature selection collapse across PII types. These results indicate that existing evaluations cannot verify whether interventions target causal pathways of PII leakage. A 9,000-run causal mediation feasibility study corroborates this: the verifier separates candidates reliably, but correlation-selected candidates rarely pass it. We therefore propose a new research mission for NLP: causally grounded PII suppression through circuit-level localization and causal mediation analysis.
\end{abstract}

\section{Introduction}
Large Language Models (LLMs) are increasingly being deployed in domains that handle sensitive information, such as healthcare, law, and finance~\cite{chen2024survey}. However, during pretraining, LLMs can memorize Personally Identifiable Information (PII) contained in their training data, which may subsequently be exposed when the model is prompted with a crafted prefix~\cite{carlini2021extracting, carlini2023quantifying}. PII leakage has therefore emerged as a critical security and ethical concern.

Recent work has attempted to mitigate PII leakage by applying Sparse Autoencoders (SAEs) to LLM activations, identifying features hypothesized to encode PII, and intervening on them through feature ablation or vector steering~\cite{frikha2025privacyscalpel}. Following common practice in PII leakage research, these studies evaluate intervention success via reductions in PII-token log-probability~\cite{kim2023propile, nakka2024piiscope}. However, this metric cannot distinguish feature-specific suppression from a magnitude artifact: a generic effect of sufficiently large perturbations regardless of direction.

To empirically examine this limitation, we conduct a selectivity experiment comparing interventions along putatively PII-related SAE feature directions with interventions along coherence-matched random directions. Selected SAE directions did not consistently outperform matched control directions on either logprob or rank metrics, suggesting that logprob-based evaluation, as commonly practiced, cannot by itself distinguish feature-specific effects from generic activation perturbations.

These findings call for a fundamental reassessment of existing SAE-based PII suppression methods. We propose a new research mission for NLP: identifying the causal pathways of PII leakage and developing interventions grounded in them, moving beyond correlational feature-level suppression toward causally grounded approaches.

The contributions of this paper are as follows.\footnote{Code, synthetic corpora, evaluation pairs, and trained SAEs are available at \url{https://github.com/LCR2001/PII-SAE-SUPPRESSION}.}
\begin{itemize}
    \item We show, using coherence-matched random direction controls and a rank-based metric, that when log-probability is used jointly for feature selection and evaluation, as in current practice, it fails to verify feature-specificity of SAE-based PII suppression interventions across 36 configurations, and that the random-dominant pattern persists against control directions drawn from the same SAE dictionary.
    \item We propose a causal research agenda for PII suppression, calling for interventions grounded in identified causal pathways rather than correlational feature associations.
    \item We show that causal mediation analysis is implementable for PII suppression via a 9,000-run feasibility study, which corroborates our diagnostic and locates the bottleneck in candidate generation.
\end{itemize}

\section{SAE-based PII Suppression in LLMs}
Recent work on PII suppression has shifted from neuron-level interventions~\cite{chen2024learnable} to SAE-based feature-level interventions~\cite{frikha2025privacyscalpel} that ablate or steer features assumed to encode PII, evaluated primarily by log-probability reduction. This success criterion is not specific to one method: reductions in PII logprob or leakage rate are the shared evaluation signal across PII leakage benchmarks and defenses~\cite{kim2023propile, nakka2024piiscope, frikha2025privacyscalpel}. However, this metric cannot distinguish feature-specific suppression from a magnitude artifact: a direction-agnostic effect of any sufficiently large hidden-state perturbation, and cannot verify whether reductions reflect interventions on PII-specific causal mechanisms. This concern is compounded by growing evidence that SAE feature directions may not be reliably distinguishable from random baselines~\cite{korznikov2026sanity, ronge2026coffee, heap2501sparse}, and more broadly that interpretability methods can fail to separate real effects from random ones~\cite{meloux2025dead}, motivating our selectivity experiment comparing selected feature interventions against coherence-matched random direction controls (additional background in Appendix~\ref{app:related_work}).

\begin{figure*}[t]
\centering
\includegraphics[width=\textwidth]{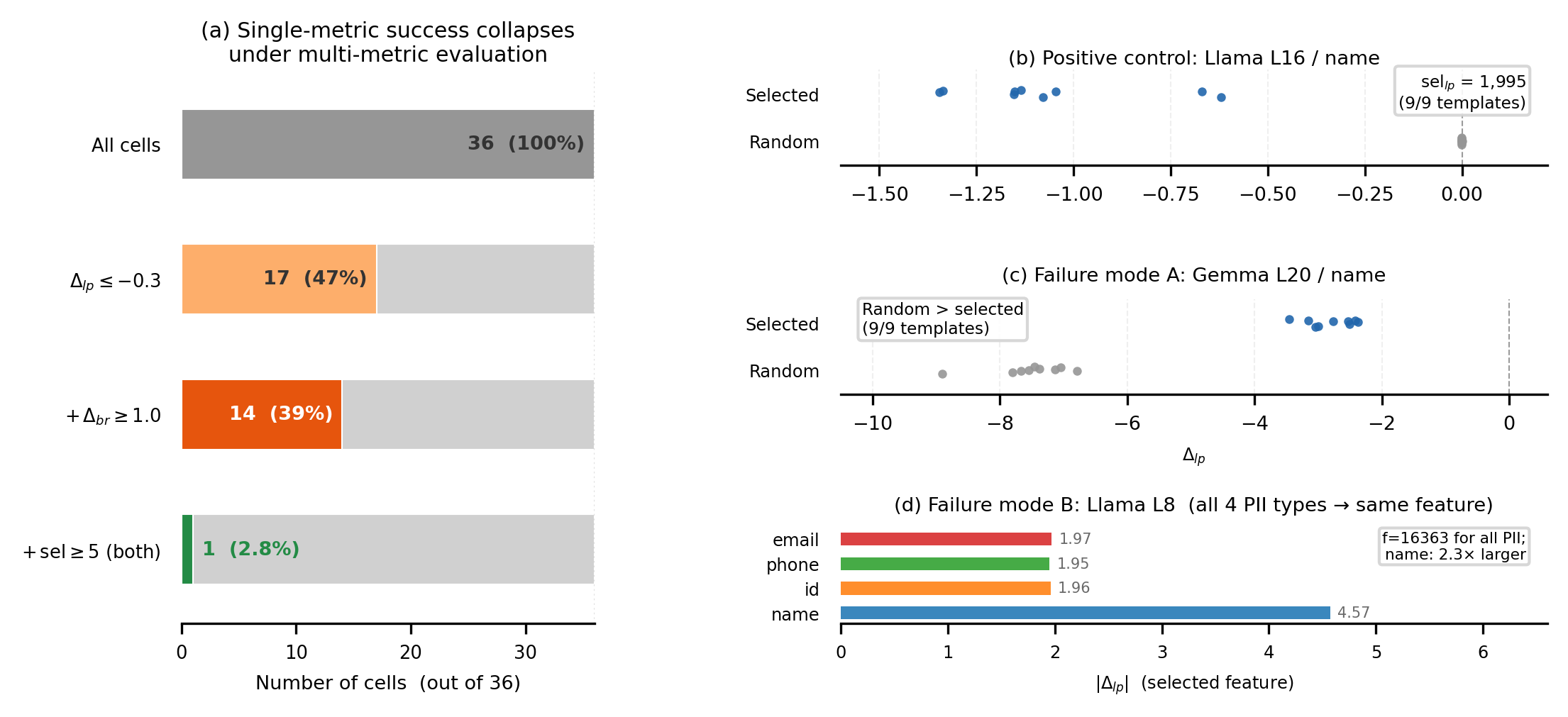}
\caption{Main diagnostic results across 36 (model, layer, PII type) 
cells. (a) Number of cells passing single-metric vs.\ multi-metric 
criteria. (b, c) Per-template $\Delta_{lp}$ for selected feature 
vs.\ coherence-matched random control. (d) $|\Delta_{lp}|$ per 
PII type in Llama L8, where f=16363 is selected for all four types.}
\label{fig:main}
\end{figure*}

\section{Experimental Setup}
\label{sec:setup}

\paragraph{Models and data.}
We evaluate SAE-based PII suppression on three general-purpose LLMs: two 8B-class models (Llama-3.1-8B-Instruct~\citep{llama3herd2024} and Qwen3-8B~\citep{qwen3technical2025}) and one 9B-class model (Gemma-2-9B~\citep{gemma2_2024}), all LoRA~\citep{hu2022lora} fine-tuned on the same synthetic medical PII corpus. Each example appends four PII fields (name, patient ID, phone number, and email) to the end of a clinical note. We use the medical domain because PII memorization in clinical LLMs is a well-documented privacy concern~\citep{lehman2021bert}, and synthetic data lets us study it without exposing real patient information. Feature discovery and final evaluation are performed on mutually disjoint PII-pair splits. Task-specific SAEs are trained on a separate 50k synthetic clinical note corpus with no entity-level overlap with either the LLM fine-tuning corpus or the evaluation pairs, so the SAEs learn general representation patterns of PII rather than memorizing specific PII entities.

\paragraph{Task-specific SAEs and feature discovery.}
For each model, we train task-specific top-$k$ SAEs~\citep{gao2025scaling} at three depth-proportional layers, approximately corresponding to 25\%, 50\%, and 75\% of the model depth (expansion factor $\times 8$, sparsity $k=32$). This depth-proportional sampling is a pre-specified design choice following prior findings that linguistic and semantic representations vary across depth \citep{tenney2019, jawahar2019, jin2025}. For each (model, layer, PII type), we identify high-activation non-dead SAE features near the PII marker as candidates and designate as the selected feature the one that yields the largest reduction in gold PII span logprob on the discovery split. This logprob-driven selection mirrors the standard approach in existing SAE-based PII suppression work that we evaluate; our contribution is to diagnose how such selected features behave under metrics beyond logprob and under random-direction controls. We note that this design is itself partially circular, since logprob serves both as the selection criterion and as the evaluated metric; Section~\ref{sec:feasibility} therefore revisits the same question with candidates selected independently of logprob.

\paragraph{Hidden-space intervention and random control.}
We use a hidden-space intervention that directly modifies the LLM hidden state along the decoder direction of a selected SAE feature. At token position $t$ in layer $\ell$:
\[
h'_t = h_t + \alpha \cdot z_t[f] \cdot W_{\mathrm{dec}}[f],
\]
where $z_t[f]$ is the SAE activation of selected feature $f$ and $W_{\mathrm{dec}}[f]$ is its decoder direction. We fix $\alpha=-2$ (sensitivity in Appendix~\ref{app:alpha_sensitivity}). Under a linear decoder, this is equivalent to a z-space edit up to a reconstruction-error term that cancels in the selectivity ratio (Table~\ref{tab:recon_diagnostics}). Due to top-$k$ sparsity, the intervention term automatically vanishes at positions where $z_t[f]=0$, so the intervention only acts where the selected feature is active without requiring an explicit PII-span mask. To distinguish feature-specific intervention from generic magnitude perturbation, we use a coherence-matched random control that preserves the token-wise magnitude $m_t = \|z_t[f] \cdot W_{\mathrm{dec}}[f]\|_2$ but replaces only the direction with an isotropic random unit vector $r$: $h'_t = h_t + \alpha \cdot m_t \cdot r$. The same $r$ is shared across all intervened positions within a forward pass, matching the span-level direction coherence of the selected-feature intervention. Random-control results are averaged over three independent seeds. Because an isotropic direction is almost surely nearly orthogonal to the local activation manifold, we additionally use a dictionary-aligned control that replaces $r$ with the normalized decoder direction of another randomly sampled non-dead feature $g$ from the same SAE, $h'_t = h_t + \alpha \cdot m_t \cdot W_{\mathrm{dec}}[g]/\|W_{\mathrm{dec}}[g]\|_2$, keeping the positions and token-wise magnitudes determined by $f$ (Appendix~\ref{app:sae_decoder_control}).

\paragraph{Evaluation metrics.}
We define a cell as one (model, layer, PII type) combination and evaluate each intervention with two signed metrics: (i) $\Delta_{\mathrm{lp}}$, the change in the mean log-probability of the gold PII span ($\Delta_{\mathrm{lp}}<0$ indicates reduced leakage), and (ii) $\Delta_{\mathrm{br}}$, the change in a rank-aware extraction-cost proxy $b_{\mathrm{rank}}(y) = \sum_{t \in y} \log_2(\mathrm{rank}_t)$ ($\Delta_{\mathrm{br}}>0$ indicates that gold PII tokens move backward in the candidate ranking and become harder to extract). Selectivity is defined as the effect-magnitude ratio between the selected feature and the random control: $\mathrm{Sel}_{\mathrm{lp}} = |\Delta_{\mathrm{lp}}^{\mathrm{target}}| / |\Delta_{\mathrm{lp}}^{\mathrm{random}}|$, and $\mathrm{Sel}_{\mathrm{br}}$ is defined analogously. We call a cell clean-strong if it satisfies all four conjunctive criteria: $\Delta_{\mathrm{lp}} \le -0.3$, $\Delta_{\mathrm{br}} \ge +1.0$, $\mathrm{Sel}_{\mathrm{lp}} \ge 5$, and $\mathrm{Sel}_{\mathrm{br}} \ge 5$. Evaluation is performed across nine prompt templates (T0: the 4-PII training format; T1--T8: isolated target PII with varying label formats including Korean, JSON, and narrative forms), and each cell-level value is averaged over $50$ pairs $\times$ $9$ templates $=$ $450$ evaluation contexts. Detailed hyperparameters, the full prompt template list, and exact definitions of category labels (clean-weak, anti-suppression, inactive, etc.) are provided in Appendix~\ref{app:experimental_details} and~\ref{app:method_details}.

\section{Results}
\label{sec:results}

Across 36 cells, only one satisfies our multi-metric criterion; the rest exhibit two non-exclusive failure modes, A (random-dominant) and B (feature selection collapse). Figure~\ref{fig:main} summarizes the findings; full results are in Appendix~\ref{app:full_results}.

\paragraph{Single-metric evaluation overestimates suppression.}
A logprob-only criterion ($\Delta_{\mathrm{lp}} \le -0.3$) labels 17/36 cells (47\%) as apparent PII suppression candidates. However, when we apply the four conjunctive criteria that jointly require a rank-aware metric and coherence-matched random-control selectivity, only one cell remains clean-strong (1/36, 2.8\%; Llama L16/name; Figure~\ref{fig:main}a). This conclusion is robust to the selectivity threshold: relaxing it to $\ge 2$ leaves only 3/36 cells (8.3\%), while tightening it to $\ge 10$ still leaves the same single cell (Appendix~\ref{app:threshold_sensitivity}).

\paragraph{Positive control: Llama L16/name.}

Llama L16/name, the only clean-strong configuration, serves as a 
positive control showing that our criterion is not impossibly strict 
(Figure~\ref{fig:main}b). It satisfies the criterion across all nine 
templates, maintaining $\mathrm{Sel}_{\mathrm{lp}} > 400$ even on
OOD Korean and JSON templates. The cell mean reaches $\mathrm{Sel}_{\mathrm{lp}} 
\approx 1{,}995$, with random-control $|\Delta_{\mathrm{lp}}|$ near 
the measurement floor ($\approx 10^{-3}$). The strict criterion 
thus admits genuine feature-specific intervention when it exists
(Appendix~\ref{app:per_template}). The same cell is the only evaluated configuration whose downstream utility is preserved: accuracy on PubMedQA, MedQA, and MMLU changes by at most one point after intervention, whereas the non-clean-strong cells Qwen L18/name and Gemma L20/name lose 12--34 points (Appendix~\ref{app:utility}). Against 20 decoder directions sampled from the same SAE, none produces a comparable effect (0/20; Appendix~\ref{app:sae_decoder_control}).

\paragraph{Failure mode A: random-dominant non-selective effect.}
Gemma L20 illustrates this pattern. Cell means fall below 
$\mathrm{Sel}_{\mathrm{lp}}=1$ for all four PII types (name 0.37, 
id 0.14, phone 0.27, email 0.09; Figure~\ref{fig:main}c). For 
Gemma L20/name, $|\Delta_{\mathrm{lp}}^{\mathrm{random}}| > 
|\Delta_{\mathrm{lp}}^{\mathrm{target}}|$ in 50/50 validation pairs 
and in 97.6\% of 450 per-context observations; the same ordering 
holds on the discovery split ($\mathrm{Sel}_{\mathrm{lp}}=0.25$). 
This is not an artifact of the isotropic control lying off the
activation manifold. All 20 dictionary-aligned directions
also produce stronger suppression than the selected direction
(20/20; Appendices~\ref{app:pattern_a_diag} and~\ref{app:sae_decoder_control}). In such random-dominant configurations, the logprob-selected SAE direction
is thus not unusually effective relative to magnitude-matched
decoder directions sampled from the same learned SAE dictionary.

\paragraph{Failure mode B: feature selection collapse.}
In 3 of 9 (model, layer) cells (Gemma L20, Llama L8, and Qwen L9) 
the discovery procedure selects the same top-1 feature for all four 
PII types, and none of these collapse cells passes the clean-strong 
criterion (Appendix~\ref{app:selected_feature_inventory}). In the 
representative case Llama L8, the selected feature has similar SAE 
activation magnitudes across PII types ($\le 5\%$ variation: name 
6.97, id 7.04, phone 6.77, email 7.01), yet its intervention effect 
is not type-agnostic: $|\Delta_{\mathrm{lp}}^{\mathrm{name}}|=4.57$ 
is about 2.3 times larger than the average for id, phone, and email 
($\approx 1.96$). Under matched activation magnitudes, this 2.3$\times$ asymmetry is
inconsistent with the feature being a generic PII-region marker,
and the same asymmetry holds in the other two collapse cells
(Gemma L20: 6.8$\times$, Qwen L9: up to 12$\times$;
Appendix~\ref{app:collapse_asymmetry}). Collapse cells also fail
the dictionary-aligned control: for Llama L8/name, 14 of 20 random
decoder directions exceed the selected direction, even though the
selected direction had dominated the isotropic control
(Appendix~\ref{app:sae_decoder_control}). Thus, shared feature
selection does not imply a shared causal intervention effect.

\section{A New Research Mission: Causally Grounded PII Suppression}
\label{sec:research_mission}
Our selectivity results expose a limitation of the evaluation framework rather than of SAEs themselves. Because selected SAE directions do not consistently outperform matched control, logprob reduction, the primary success criterion in prior work, cannot confirm whether an intervention targets a PII-specific causal mechanism or merely perturbs activations. From the perspective of Pearl's ladder of causation~\cite{pearl2009causality, pearl2018book}, current practices remain at the level of association rather than intervention. We therefore propose two directions toward causally grounded PII suppression and report a feasibility study.

The first direction is to identify the causal pathways underlying PII leakage through circuit-level localization. \citet{lasy2025understanding} show that memorization initiation and maintenance in LLMs can be supported by distinct circuits. Applying circuit localization methods such as path patching~\cite{wang2022interpretability} to PII leakage could identify which attention heads and MLP layers causally contribute to PII generation.

The second direction is to adopt causal mediation analysis, introduced to NLP by \citet{vig2020investigating}, for both identifying genuine causal mediators of PII leakage and verifying intervention effects: paired prompts with ($x_1$) and without ($x_0$) the PII are constructed, and the Natural Indirect Effect (NIE) of an SAE feature $f$ measures the extent to which $f$ mediates PII generation. Applying this to PII is not straightforward, since a clean counterfactual must be defined and effects must be aggregated over a multi-token span; the feasibility study below confronts both issues.

\subsection{Feasibility Study}
\label{sec:feasibility}
To verify that this framework is implementable rather than merely proposable, we conduct a feasibility study on the same three models. For each (model, layer, PII type), we construct 50 pairs of text containing the original PII ($x_1$) and counterfactual text ($x_0$) in which the PII is replaced with a different value, and write $\mathrm{lp}(y \mid x)$ for the log-probability of the original PII span $y$ given text $x$. The intervention position is the token immediately preceding the PII span. Because $x_0$ and $x_1$ share the preceding text, standard activation patching at this position yields an identically zero NIE (Appendix~\ref{app:mediation_study}). We therefore use a noising-style intervention that zero-ablates feature $f$ during the $x_1$ run, which conceptually corresponds to the noising intervention of causal tracing~\cite{meng2022locating}. We define $\mathrm{TE} = \mathrm{lp}(y \mid x_1) - \mathrm{lp}(y \mid x_0)$ and $\mathrm{NIE}_f = \mathrm{lp}(y \mid x_1, f{=}0) - \mathrm{lp}(y \mid x_0)$, and the mediation fraction as $1 - \mathrm{mean}(\mathrm{NIE}_f)/\mathrm{mean}(\mathrm{TE})$, computed as a ratio of effects averaged across pairs. The control condition $\mathrm{NIE}_{\mathrm{rand}}$ uses a random direction of matched magnitude. Unlike the logprob-driven selection of Section~\ref{sec:setup}, candidate features are selected independently based on activation rate at the position immediately preceding the PII, so that logprob is not used for both feature selection and evaluation. In total, 180 cells $\times$ 50 pairs yield 9,000 causal patching runs (Appendix~\ref{app:mediation_study}). Table~\ref{tab:mediation_study} summarizes the results.

\begin{table}[t]
\centering
\footnotesize
\setlength{\tabcolsep}{2pt}
\begin{tabular}{llrrrrr}
\toprule
Model & PII & TE & $\mathrm{NIE}_f$ & $\mathrm{NIE}_{\mathrm{rand}}$ & Ratio & Indist. \\
\midrule
Gemma        & email & $+2.580$ & $2.504$  & $2.574$  & 0.973 & 100.0 \\
             & id    & $-0.413$ & $-0.420$ & $-0.399$ & 1.053 & 100.0 \\
             & name  & $+0.609$ & $0.408$  & $0.649$  & \textbf{0.628} & \textbf{73.3} \\
\midrule
Llama        & email & $+3.560$ & $3.558$  & $3.559$  & 1.000 & 100.0 \\
             & id    & $-0.233$ & $-0.241$ & $-0.237$ & 1.016 & 100.0 \\
             & name  & $-0.283$ & $-0.290$ & $-0.283$ & 1.023 & 100.0 \\
             & phone & $+0.067$ & $0.064$  & $0.066$  & 0.959 & 100.0 \\
\midrule
Qwen         & email & $+2.466$ & $2.464$  & $2.466$  & 0.999 & 100.0 \\
             & name  & $-0.604$ & $-0.658$ & $-0.626$ & 1.051 & 100.0 \\
             & phone & $-0.065$ & $-0.068$ & $-0.068$ & 1.000 & 100.0 \\
\midrule
\multicolumn{6}{l}{Overall (150 features, $|\mathrm{TE}| \ge 0.05$)} & 97.3 \\
\bottomrule
\end{tabular}
\caption{Causal mediation feasibility for (model, PII type) combinations with $|\mathrm{TE}| \geq 0.05$ (150 of 180 candidate features). NIE$_f$ and NIE$_{\mathrm{rand}}$ are means over 15 candidate features; Ratio is their ratio, and Indist.\ is the percentage of candidates whose individual ratio lies in $(0.5, 2.0)$. Bold marks the only cell in which some candidates are distinguishable from random; it is also the only non-email cell with a positive memorization signal. Full protocol in Appendix~\ref{app:mediation_study}.}
\label{tab:mediation_study}
\end{table}

First, memorization is PII-type specific. Email shows a clear memorization signal ($\mathrm{TE} > 2.4$) across all three models, phone number and patient ID show largely no memorization, and name is inconsistent across models ($+0.61$ Gemma, $-0.28$ Llama, $-0.60$ Qwen). Second, correlation-selected features are indistinguishable from the random control, including in the email cells where memorization is strongest. Of the 150 candidate features in cells with $|\mathrm{TE}| \ge 0.05$, 146 (97.3\%) have $\mathrm{NIE}_f / \mathrm{NIE}_{\mathrm{rand}}$ within $(0.5, 2.0)$. All candidates are indistinguishable from random in 9 of 10 (model, PII type) combinations. The sole exception, Gemma-2-9B/name (11/15), is precisely the one combination exhibiting a positive memorization signal. This corroborates the finding of Section~\ref{sec:results} under a fully independent causal methodology. Third, the logprob-selected features of Section~\ref{sec:setup} and the activation-rate candidates share no feature across all 36 combinations, yet both pools show a similarly low rate of passing causal verification (Appendix~\ref{app:mediation_study}).

Taken together, causal mediation analysis performs correctly, separating candidates by whether genuine memorization is present; the bottleneck is candidate generation, since neither correlation-based signal guarantees causal specificity. Our multi-metric diagnostic is therefore necessary but not sufficient, and circuit-level localization is needed to supply causally motivated candidates.

\section{Conclusion}
\label{sec:conclusion}
Across 36 (model, layer, PII type) configurations, we show that logprob reduction, the standard evaluation signal in SAE-based PII suppression, does not distinguish feature-specific intervention from generic magnitude perturbation. Only one cell (Llama L16/name) passes both the coherence-matched random control and the rank-aware metric. This confirms that our evaluation protocol can identify genuine feature-specific effects when they exist, while showing that, in the remaining 35 cells, conventional evaluation cannot verify whether the selected feature is causally involved in PII leakage.
Multi-metric evaluation is a diagnostic tool, not a final solution, and the fundamental goal is to identify the causal pathways underlying PII leakage. We propose causally grounded PII suppression, through circuit-level localization and causal mediation analysis, as a new research mission for NLP. A feasibility study confirms that mediation analysis reliably separates cases with and without memorization, while correlation-selected candidates rarely pass it. The open problem is therefore causally motivated candidate generation.

\section*{Limitations}

This study has several limitations.

\paragraph{Model scale and domain.}
Our experiments are limited to 8B/9B-class general-purpose LLMs. We do not test whether the same patterns hold for substantially larger models, closed-source models, or domain-specialized LLMs, mainly because task-specific SAE analysis requires collecting large-scale activations and training separate SAEs for each model and layer. In particular, domain-specialized models may exhibit different activation distributions due to their pretraining and fine-tuning corpora; developing SAE training protocols suitable for such models, and testing whether our findings replicate on them, remain future work; our released SAE training and evaluation code is intended to lower the cost of such replication.

\paragraph{Hidden-space vs.\ z-space intervention.}
Our intervention preserves the SAE reconstruction-error term $(\mathbf{h}_t - \hat{\mathbf{h}}_t)$ that a full encode-decode pipeline (as in PrivacyScalpel) would discard. While the selectivity ratio is invariant to this error term, the absolute magnitudes of $\Delta_{lp}$ and $\Delta_{br}$ may shift under z-space intervention. Direct z-space replication is left to future work.

\paragraph{SAE capacity.}
We evaluate task-specific SAEs with expansion factor $\times 8$ and fixed sparsity $k=32$. We do not ablate larger expansion factors such as $\times 16$ or $\times 32$, alternative sparsity levels, or different SAE objectives. Since SAE capacity may affect feature isolation, future work should test whether the failure modes identified here, especially feature selection collapse (Failure mode B) and random-dominant non-selective effects (Failure mode A), persist under larger or differently trained SAEs.

\paragraph{Synthetic data.}
Our evaluation uses a synthetic medical PII corpus. We avoid real patient records due to ethical and regulatory constraints. Synthetic data allows us to control PII values, clinical context, and entity-level disjoint splits, but it cannot fully capture the linguistic diversity, formatting noise, or annotation ambiguity of real medical records. Whether the same patterns hold in real-world private corpora remains open. At the same time, our core claim is methodological rather than dataset-specific: logprob-only evaluation can fail to distinguish feature-specific interventions from generic magnitude perturbations.

\paragraph{Selective layer coverage.}
We train and evaluate SAEs at three pre-specified depth-proportional layers, approximately 25\%, 50\%, and 75\% of each model's depth. This design avoids choosing layers based on downstream intervention outcomes, but it is not an exhaustive layer-wise search. Other layers may contain stronger feature-specific intervention directions, or may exhibit different failure modes. Exhaustive layer-wise SAE analysis is left for future work.

\paragraph{Circular selection design.}
The main experiments select features by logprob reduction and then evaluate them by logprob, mirroring current practice but making the design partially circular. Our conclusion is accordingly scoped to this joint use. The feasibility study (Section~\ref{sec:feasibility}) selects candidates by activation rate instead and reaches the same conclusion, but other selection criteria, such as monosemanticity or top-activating examples, remain untested.

\paragraph{Single-feature intervention.}
Our experiments focus on single-feature hidden-space interventions using one selected SAE feature. Thus, when a selected feature is indistinguishable from random controls or produces anti-suppression, this does not rule out multi-feature or circuit-level interventions. Rather, our results suggest that PII leakage may arise through causal pathways more complex than a single SAE feature; identifying and intervening on such pathways remains a key direction for future work.

\paragraph{Extraction risk and utility.}
We do not evaluate full adversarial extraction attacks. Our metrics (gold PII span logprob, a rank-aware extraction-cost proxy, and random-control selectivity) are mechanism-specific diagnostics, not privacy guarantees against arbitrary attackers. Extraction success and feature-specificity are related but distinct questions: extraction rate can fall because an intervention degrades generation in general rather than because it disrupts a PII-specific pathway, as the utility losses of Qwen L18/name and Gemma L20/name illustrate (Appendix~\ref{app:utility}). Adversarial extraction benchmarks should therefore be seen as a complementary end-to-end evaluation to be run alongside our diagnostics, not as a substitute for them. Our utility evaluation itself covers only three configurations and three benchmarks.

\section*{Ethical considerations}
This work studies SAE-based PII suppression in LLMs, motivated by privacy concerns surrounding LLM memorization of sensitive information. To avoid exposing real patient data, all PII used in this study is fully synthetic. The synthetic medical PII corpus and clinical notes were generated for the controlled study of intervention behavior and do not contain any real-world identifying information.
Our results highlight that prior SAE-based PII suppression evaluations may overestimate genuine privacy protection. While this finding could inform improved privacy defenses, the diagnostic methodology operates at the level of evaluation and does not enable new attack capabilities beyond what is already accessible to any black-box adversary. We encourage future work to combine the multi-metric diagnostic framework with extraction-attack benchmarks to verify substantive privacy guarantees.

\section*{Acknowledgments}
This work was supported by Korea Internet \& Security Agency(KISA) grant funded by the Korea government(PIPC) (No.RS-2026-25526342, Development of Technologies for Preventing Sensitive Information Inference and Risk Assessment in Foundation Model Operations).

\bibliography{custom}

\appendix

\section{Appendix}

\subsection{Additional Background on PII Suppression}
\label{app:related_work}

\subsubsection{PII Memorization in LLMs}
LLMs can memorize PII contained in their pretraining data, which may then be leaked through verbatim reproduction when the model is prompted with a particular prefix. \citet{carlini2021extracting} demonstrated that verbatim sequences containing PII, including names, phone numbers, and email addresses, can be extracted from GPT-2 using only black-box queries, and showed that memorization becomes more severe as model size increases \cite{carlini2023quantifying}. \citet{ippolito2022preventing} further showed that even a Bloom-filter-based defense that perfectly prevents verbatim memorization can be bypassed by minimally modified style-transfer prompts, allowing memorized information to still be leaked. These findings warn that defenses satisfying surface-level metrics do not necessarily guarantee substantive privacy protection, closely aligning with the limitations of evaluation protocols discussed in the main section.

\subsubsection{SAE-based PII Suppression}
With advances in interpretability research on the internal representations of LLMs, studies on PII suppression has increasingly shifted toward interventions on model internals. \citet{chen2024learnable} combined learnable binary weight masks with adversarial training to localize and deactivate privacy neurons associated with PII memorization, aiming to suppress PII leakage. However, such neuron-level approaches are limited in both interpretability and precision due to polysemanticity, as individual neurons may represent multiple concepts simultaneously. Therefore, SAE-based approaches have been proposed as a more interpretable alternative. By decomposing LLM activations into high-dimensional sparse feature spaces, SAEs aim to learn monosemantic representations in which individual features correspond to specific concepts~\cite{bricken2023monosemanticity, cunningham2023sparse}. PrivacyScalpel~\cite{frikha2025privacyscalpel} is a representative study that applies SAEs to PII suppression. It first identifies the layer where PII representations are most concentrated through feature probing, trains a k-SAE on the activations of that layer to isolate privacy-sensitive features in a monosemantic space, and then suppresses PII-related features through feature ablation and vector steering. On Gemma2-2B and Llama2-7B, PrivacyScalpel reportedly reduces the email leakage rate from 5.15\% to 0.0\% while preserving more than 99.4\% of the original model performance, demonstrating a stronger privacy-utility trade-off compared with neuron-level interventions. However, its evaluation does not determine whether the reduction in email leakage results from selective suppression of PII-sensitive features or from a generic activation perturbation that degrades the model’s ability to generate email-formatted strings.

\subsubsection{Critiques of SAE Feature Reliability}
Alongside the rapid adoption of SAE-based interpretability research, recent work has increasingly raised critical questions about the substantive meaning and reliability of SAE features. \citet{ronge2026coffee} stress-test  Anthropic-style SAE-based feature steering using Llama 3.1 and find that steering effects exhibit substantial fragility depending on layer selection, steering magnitude, and context. These findings suggest that SAE feature directions may function only under limited conditions, casting doubt on whether they provide stable and generalizable causal effects. Furthermore, \citet{korznikov2026sanity} systematize this concern by showing that random-feature baselines can perform comparably to fully trained SAEs across interpretability, sparse probing, and causal editing evaluations, raising a fundamental question about whether SAE features provide meaningful information. Our paper extends this critique to the privacy domain by examining it in the concrete setting of PII suppression. Specifically, we use coherence-matched random controls to directly test whether evaluation metrics used in prior PII suppression studies, such as email leakage rate, can distinguish feature-specific interventions from generic activation perturbations. In doing so, we instantiate \citet{korznikov2026sanity}’s general critique as a methodological critique of evaluation practices in SAE-based PII suppression. Related sanity-check work shows that SAEs trained on randomly initialized transformers yield features that look comparably interpretable~\cite{heap2501sparse}, and that interpretability pipelines more generally can report effects that are indistinguishable from chance~\cite{meloux2025dead}; our dictionary-aligned control (Appendix~\ref{app:sae_decoder_control}) is designed in this spirit.

\subsection{Models, Data, and SAE Training}
\label{app:experimental_details}
All experiments, including SAE training and intervention evaluation, were conducted on a single NVIDIA RTX 4090 (24 GB VRAM).

\paragraph{Models and data.}
We use three LoRA fine-tuned LLMs: two 8B-class models (Llama-3.1-8B-Instruct and Qwen3-8B) and one 9B-class model (Gemma-2-9B), all fine-tuned on the same synthetic medical PII corpus. These publicly available models are used under their respective licenses: Llama-3.1-8B-Instruct under the Llama 3.1 Community License (Meta Platforms, Inc.), Qwen3-8B under the Apache 2.0 License (Alibaba Cloud), Gemma-2-9B under the Gemma Terms of Use (Google). Each example appends four PII fields (name, patient ID, phone number, and email) to the end of a clinical note. Table~\ref{tab:model_layers} summarizes the hidden dimension, total number of layers, and the task-specific SAE training layers for each model.

\begin{table}[h]
\centering
\small
\begin{tabular}{lccc}
\hline
Model & Depth & $d_{\mathrm{model}}$ & SAE layers \\
\hline
Gemma-2-9B & 42 & 3584 & 9, 20, 31 \\
Llama-3.1-8B-Instruct & 32 & 4096 & 8, 16, 24 \\
Qwen3-8B & 36 & 4096 & 9, 18, 27 \\
\hline
\end{tabular}
\caption{Model and layer configuration. SAE layers are selected at approximately 25\%, 50\%, 75\% depth.}
\label{tab:model_layers}
\end{table}

\paragraph{Synthetic corpus generation.}
All clinical notes and PII values are generated by deterministic templates without any LLM. A note is assembled from fixed sentence pools for chief complaint, history of present illness, past medical history, current medications, physical examination, and assessment/plan, with age, gender, symptom duration, and severity drawn at random, followed by a handover block listing the four PII fields (\texttt{Patient Name}, \texttt{Patient ID}, \texttt{Phone}, \texttt{Email}); this block is the T0 training format in Appendix~\ref{app:templates}. Names are drawn from a fixed first-name~$\times$~last-name pool, patient IDs are sequential (\texttt{PID-00001}, \ldots), phone numbers are deterministic arithmetic functions of the entity index in North American format, and email addresses are derived from the name with an optional numeric tag and one of seven common domains.

The evaluation registry contains 600 entities (seed 42; 30~first names~$\times$~20~last names, enumerated exhaustively). The SAE-training corpus contains 50,000 entities generated with a separate seed (1234), larger sentence pools, a wider age range, and first- and last-name pools from which every name in the evaluation registry has been removed; patient IDs start at \texttt{PID-10001}, and phone numbers and emails are rejection-sampled against the evaluation registry, so no PII value or name is shared between the two. The LoRA fine-tuning corpus uses a third, disjoint name pool; we verified programmatically that no name, patient ID, phone number, email, or note prefix overlaps across the three splits. Within the evaluation registry, feature discovery uses 10 pairs (indices 0--9) and all reported evaluation uses a disjoint set of 50 validation pairs (indices 300--349). The synthetic corpus, evaluation pairs, prompt templates, and the generation scripts that produce them are released at the repository linked in footnote~1.
For evaluation, we use a total of 600 PII pairs, with feature discovery and final evaluation performed on disjoint splits (Table~\ref{tab:data_splits}).

\begin{table}[h]
\centering
\small
\resizebox{\columnwidth}{!}{
\begin{tabular}{lcc}
\hline
Split & Pair indices & Purpose \\
\hline
Discovery & 0--9 & Feature discovery \\
Diagnostic & 0--19 & Reconstruction diagnostics \\
Validation & 300--349 & Main multi-metric evaluation \\
SAE training & separate 50k corpus & Task-SAE training \\
\hline
\end{tabular}
}
\caption{Data split used for feature discovery, diagnostics, and final evaluation. The SAE training corpus is entity-level disjoint from the 600 evaluation pairs.}
\label{tab:data_splits}
\end{table}

\paragraph{SAE training.}
For each model, we train top-$k$ sparse autoencoders at three depth-proportional layers, corresponding to approximately 25\%, 50\%, and 75\% of the model depth. This design follows prior work that compares transformer representations at fixed relative depths \citep{jin2025}, as well as studies showing that linguistic and task-relevant information varies across layers \citep{tenney2019, jawahar2019}. The sampled layers are not chosen based on SAE quality metrics or downstream intervention performance; thus, any layer-dependent differences observed in our experiments should be interpreted as differences across model depth rather than as outcome-driven selection bias. All SAEs are trained with the same hyperparameters (Table~\ref{tab:sae_hparams}).

\begin{table}[h]
\centering
\small
\begin{tabular}{lc}
\hline
Hyperparameter & Value \\
\hline
Expansion factor & $\times 8$ \\
Sparsity $k$ & 32 \\
Auxiliary loss weight & 0.25 \\
Dead feature threshold & 1000 inactive steps \\
Training steps & 80{,}000 \\
Batch size & 4{,}096 tokens \\
Learning rate & $3 \times 10^{-4}$ \\
\hline
\end{tabular}
\caption{Task-specific SAE training hyperparameters.}
\label{tab:sae_hparams}
\end{table}

All trained SAEs achieve over 98.8\% variance explained on the held-out SAE training subset. However, corpus-level reconstruction quality does not necessarily reflect quality at intervention-relevant PII token positions, so we separately report PII-position reconstruction error in Table~\ref{tab:recon_diagnostics}.

\begin{table}[h]
\centering
\small
\resizebox{\columnwidth}{!}{
\begin{tabular}{lcccc}
\hline
Model & Layer & Rel. err. all & Rel. err. PII & Cosine PII \\
\hline
Gemma-2-9B             & 9  & 5.11\%  & 2.78\%  & 0.9996 \\
Gemma-2-9B             & 20 & 10.28\% & 6.38\%  & 0.9980 \\
Gemma-2-9B             & 31 & 9.74\%  & 8.82\%  & 0.9961 \\
\hline
Llama-3.1-8B-Instruct  & 8  & 5.20\%  & 1.99\%  & 0.9998 \\
Llama-3.1-8B-Instruct  & 16 & 6.77\%  & 13.53\% & 0.9907 \\
Llama-3.1-8B-Instruct  & 24 & 10.45\% & 39.20\% & 0.9192 \\
\hline
Qwen3-8B               & 9  & 5.88\%  & 2.11\%  & 0.9998 \\
Qwen3-8B               & 18 & 8.60\%  & 2.62\%  & 0.9997 \\
Qwen3-8B               & 27 & 11.28\% & 35.59\% & 0.9337 \\
\hline
\end{tabular}
}
\caption{SAE reconstruction quality across nine (model, layer) cells. Relative error is $\|\mathbf{h} - \hat{\mathbf{h}}\|/\|\mathbf{h}\|$; "all" averages over all token positions, "PII" over PII token positions only. Cosine similarity is computed at PII token positions. Deep layers (Llama L24, Qwen L27) show higher PII-position reconstruction error.}
\label{tab:recon_diagnostics}
\end{table}

\subsection{Prompt Templates}
\label{app:templates}

To evaluate cross-template robustness, we use nine prompt templates. T0 follows the original training format, while T1--T8 introduce variations including different label lengths, natural language phrasing, Korean labels, JSON format, and narrative format (Table~\ref{tab:templates}).

\begin{table}[h]
\centering
\small
\setlength{\tabcolsep}{4pt}
\begin{tabular}{lll}
\hline
ID & Example format for name PII & Context \\
\hline
T0 & \texttt{Patient Name: \{name\}} & all 4 PII fields \\
T1 & \texttt{Patient Name: \{name\}} & target PII only \\
T2 & \texttt{Name: \{name\}} & target PII only \\
T3 & \texttt{Full name: \{name\}} & target PII only \\
T4 & \texttt{The patient's name is \{name\}} & target PII only \\
T5 & \texttt{환자 이름: \{name\}} & target PII only \\
T6 & \texttt{Patient: \{name\}} & target PII only \\
T7 & \texttt{\{"name": "\{name\}"\}} & target PII only \\
T8 & \texttt{The patient, \{name\}, ...} & target PII only \\
\hline
\end{tabular}
\caption{Prompt templates used for cross-template evaluation. T0 retains the original training prompt that contains all four PII fields (name, patient ID, phone, email) together; T1--T8 isolate the target PII field only, with varying label formats. T8 is abbreviated (full text: ``The patient, \{name\}, presented with chest pain.''). The same template family is adapted to patient ID, phone number, and email PII.}
\label{tab:templates}
\end{table}
\subsection{Feature Discovery, Intervention, and Metrics}
\label{app:method_details}

\paragraph{Feature discovery.}
Feature discovery is performed only on the discovery split. We first extract hidden states near the PII marker and encode them with the SAE to obtain sparse activations $z$. For each PII type, we select non-dead features with large mean activation magnitude as candidates, where a dead feature is one that has not been activated for a fixed inactivity window by the end of SAE training. For each candidate, we apply hidden-space intervention directly and designate as the selected feature for each (model, layer, PII type) the one that yields the largest reduction in gold PII span logprob on the discovery split. The selected feature is fixed before validation, and final results are reported only on the held-out validation split.

\paragraph{Hidden-space intervention.}
Our main intervention operates in hidden space rather than in SAE latent space. Instead of modifying the SAE latent and reconstructing the hidden state through the decoder, we directly steer the original LLM hidden state using the decoder direction of the selected SAE feature. At token position $t$ in layer $\ell$:
\[
h'_t = h_t + \alpha \cdot z_t[f] \cdot W_{\mathrm{dec}}[f],
\]
where $z_t[f]$ is the SAE activation of selected feature $f$ at token $t$, and $W_{\mathrm{dec}}[f]$ is its decoder direction. We fix $\alpha=-2$ for all main experiments and do not retune it per model, layer, or PII type. The intervention hook is applied to all token positions in the forward pass; however, because top-$k$ SAE activations are sparse, the intervention term vanishes automatically at tokens where the selected feature is inactive ($z_t[f]=0$). Thus, we do not use an explicit PII-span intervention mask, and evaluation is restricted to the gold PII span.

\paragraph{Coherence-matched random direction control.}
To distinguish feature-specific effects from generic magnitude perturbations, we compare each selected feature intervention against a coherence-matched random direction control. The random control preserves the same $\alpha$ and token-wise intervention magnitude, but replaces the SAE decoder direction with an isotropic random unit vector. For the target contribution $c_t = z_t[f] W_{\mathrm{dec}}[f]$, we compute $m_t = \|c_t\|_2$. In each forward pass, we sample one isotropic random unit vector $r$ and share it across all intervened token positions, matching the span-level direction coherence of the selected-feature intervention, which uses the same $W_{\mathrm{dec}}[f]$ at all positions:
\[
h'_t = h_t + \alpha \cdot m_t \cdot r.
\]
Thus, the selected-feature intervention and the random control share the same intervention strength, token-wise magnitude pattern, and span-level direction coherence; their intended difference is only whether the direction is the selected SAE decoder direction or an isotropic random direction. We average random-control results over three independent seeds.

\paragraph{Metrics and selectivity.}
We evaluate each intervention using two signed metrics. The first is the change in mean log-probability of the gold PII span,
\[
\Delta_{\mathrm{lp}} = \mathrm{lp}_{\mathrm{int}} - \mathrm{lp}_{\mathrm{base}},
\]
where $\Delta_{\mathrm{lp}}<0$ indicates reduced gold PII likelihood after intervention. The second is the change in a rank-aware extraction-cost proxy,
\[
b_{\mathrm{rank}}(y) = \sum_{t \in y} \log_2(\mathrm{rank}_t),
\]
with intervention-induced change $\Delta_{\mathrm{br}} = b_{\mathrm{rank,int}} - b_{\mathrm{rank,base}}$; $\Delta_{\mathrm{br}}>0$ indicates that gold PII tokens move backward in the candidate ranking, increasing the rank-based extraction cost. Selectivity is defined as the effect magnitude of the selected-feature intervention (denoted with the $^{\mathrm{target}}$ superscript) divided by the effect magnitude of the random direction control:
\[
\mathrm{Sel}_{\mathrm{lp}} = \frac{|\Delta_{\mathrm{lp}}^{\mathrm{target}}|}{|\Delta_{\mathrm{lp}}^{\mathrm{random}}|},
\quad
\mathrm{Sel}_{\mathrm{br}} = \frac{|\Delta_{\mathrm{br}}^{\mathrm{target}}|}{|\Delta_{\mathrm{br}}^{\mathrm{random}}|}.
\]
Selectivity is a magnitude-based diagnostic for feature-specificity; the directionality of suppression or anti-suppression is interpreted separately through the signs of raw $\Delta_{\mathrm{lp}}$ and $\Delta_{\mathrm{br}}$.

\paragraph{Category thresholds.}
We use threshold-based categories to summarize intervention behavior in figures and summary tables. These categories are not used for intervention tuning; they serve only as post-hoc labels for interpreting the results. Each cell is assigned to one of seven categories based on five criteria, summarized in Tables~\ref{tab:thresholds_criterion} and~\ref{tab:thresholds_category}.

\begin{table}[h]
\centering
\small
\begin{tabular}{ll}
\hline
Criterion & Meaning \\
\hline
$\Delta_{\mathrm{lp}} \leq -0.3$ & logprob suppression \\
$\Delta_{\mathrm{br}} \geq +1.0$ & rank-cost improvement \\
$\Delta_{\mathrm{lp}} > +0.1$ & logprob anti-suppression \\
$\Delta_{\mathrm{br}} < -0.3$ & rank-cost degradation \\
$\mathrm{Sel}_{\mathrm{lp}}, \mathrm{Sel}_{\mathrm{br}} \geq 5$ & strong feature-specificity \\
\hline
\end{tabular}
\caption{Five criteria used for category labels.}
\label{tab:thresholds_criterion}
\end{table}

\begin{table}[h]
\centering
\footnotesize
\setlength{\tabcolsep}{3pt}
\resizebox{\columnwidth}{!}{%
\begin{tabular}{ll}
\hline
Category & Definition \\
\hline
clean-strong & $\Delta_{\mathrm{lp}} \le -0.3$, $\Delta_{\mathrm{br}} \ge +1.0$, $\mathrm{Sel}_{\mathrm{lp}}, \mathrm{Sel}_{\mathrm{br}} \ge 5$ \\
clean-weak & $\Delta_{\mathrm{lp}} \le -0.3$, $\Delta_{\mathrm{br}} \ge +1.0$, $\mathrm{Sel} < 5$ on $\ge$1 metric \\
anti & $\Delta_{\mathrm{lp}} > +0.1$ or $\Delta_{\mathrm{br}} < -0.3$ \\
lp-only & $\Delta_{\mathrm{lp}} \le -0.3$ but $\Delta_{\mathrm{br}} < +1.0$ \\
br-only & $\Delta_{\mathrm{br}} \ge +1.0$ but $\Delta_{\mathrm{lp}} > -0.3$ \\
inactive & $z_t[f] \approx 0$ \\
noise & none of the above \\
\hline
\end{tabular}%
}
\caption{Category definitions used for post-hoc labeling. Each cell is assigned exactly one category. Raw $\Delta$ and Sel values are reported alongside category labels.}
\label{tab:thresholds_category}
\end{table}

\subsection{Full Per-Cell Results}
\label{app:full_results}

\begin{figure*}[t]
\centering
\includegraphics[width=\textwidth]{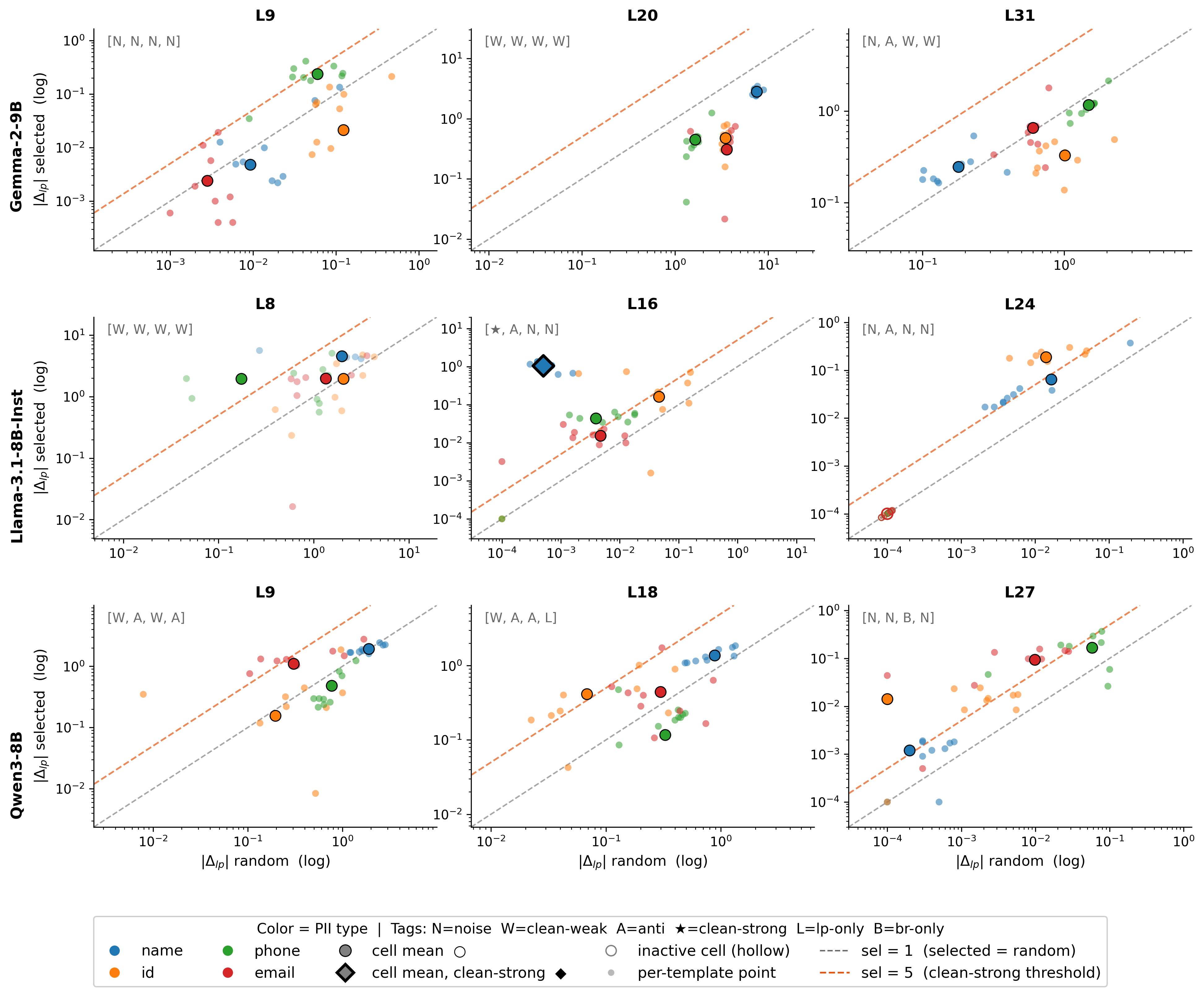}
\caption{Selected feature intervention effect ($|\Delta_{\mathrm{lp}}^{\mathrm{target}}|$, y-axis) versus coherence-matched random direction effect ($|\Delta_{\mathrm{lp}}^{\mathrm{random}}|$, x-axis), per (model, layer, PII type) configuration; both axes log scale. Large markers indicate cell mean over 9 templates $\times$ 50 validation pairs (450 evaluation contexts); small dots indicate per-template means within a cell. Gray dashed: $\mathrm{Sel}_{\mathrm{lp}}=1$ (target $=$ random). Orange dashed line: $\mathrm{Sel}_{\mathrm{lp}}=5$ (clean-strong selectivity threshold). Diamond ($\blacklozenge$): cell satisfying all clean-strong criteria; hollow markers: inactive configurations ($z_t[f]\approx 0$). Of 36 configurations, only Llama L16/name qualifies as clean-strong. Top-left tag in each subplot encodes category labels for [name, id, phone, email] (N=noise, W=clean-weak, A=anti, L=lp-only, B=br-only, $\bigstar$=clean-strong). The clean-strong label ($\blacklozenge$) reflects the full multi-metric criterion including rank-aware $\Delta_{\mathrm{br}}$ and signed-metric checks.}
\label{fig:selectivity_scatter}
\end{figure*}

The categorical analysis in Section~\ref{sec:results} is computed from cell-level means over all 36 cells. Table~\ref{tab:full_cell_results} reports, for every cell, the $\Delta_{\mathrm{lp}}$, $\Delta_{\mathrm{br}}$, both selectivity values, and the category label for the selected-feature intervention and its coherence-matched random control. All values are averaged over 50 validation pairs $\times$ 9 prompt templates $=$ 450 evaluation contexts.

\begin{table*}[h]
\centering
\small
\begin{tabular}{llrrrrrrl}
\hline
Model & Layer/PII & $\Delta_{\mathrm{lp}}^{\mathrm{T}}$ & $\Delta_{\mathrm{lp}}^{\mathrm{R}}$ & $\mathrm{Sel}_{\mathrm{lp}}$ & $\Delta_{\mathrm{br}}^{\mathrm{T}}$ & $\Delta_{\mathrm{br}}^{\mathrm{R}}$ & $\mathrm{Sel}_{\mathrm{br}}$ & Category \\
\hline
Gemma & 9/name & $-0.005$ & $+0.009$ & 0.52 & $+0.053$ & $-0.079$ & 0.68 & noise \\
Gemma & 9/id & $+0.021$ & $-0.123$ & 0.17 & $+0.111$ & $+0.310$ & 0.36 & noise \\
Gemma & 9/phone & $-0.235$ & $-0.060$ & 3.95 & $+0.144$ & $-0.040$ & 3.56 & noise \\
Gemma & 9/email & $+0.002$ & $+0.003$ & 0.88 & $+0.007$ & $+0.004$ & 1.54 & noise \\
Gemma & 20/name & $-2.805$ & $-7.524$ & 0.37 & $+4.806$ & $+12.906$ & 0.37 & clean-weak \\
Gemma & 20/id & $-0.475$ & $-3.460$ & 0.14 & $+3.166$ & $+18.171$ & 0.17 & clean-weak \\
Gemma & 20/phone & $-0.450$ & $-1.646$ & 0.27 & $+6.800$ & $+19.997$ & 0.34 & clean-weak \\
Gemma & 20/email & $-0.308$ & $-3.583$ & 0.09 & $+5.806$ & $+33.087$ & 0.17 & clean-weak \\
Gemma & 31/name & $-0.246$ & $-0.179$ & 1.38 & $-0.085$ & $+0.269$ & 0.32 & noise \\
Gemma & 31/id & $-0.329$ & $-1.006$ & 0.33 & $-0.557$ & $+4.963$ & 0.11 & anti \\
Gemma & 31/phone & $-1.168$ & $-1.486$ & 0.79 & $+6.891$ & $+7.833$ & 0.88 & clean-weak \\
Gemma & 31/email & $-0.655$ & $-0.602$ & 1.09 & $+3.034$ & $+3.279$ & 0.93 & clean-weak \\
\hline
Llama & 8/name & $-4.565$ & $-1.967$ & 2.32 & $+6.141$ & $+2.818$ & 2.18 & clean-weak \\
Llama & 8/id & $-1.958$ & $-2.047$ & 0.96 & $+7.988$ & $+7.947$ & 1.00 & clean-weak \\
Llama & 8/phone & $-1.951$ & $-0.174$ & 11.21 & $+14.422$ & $+4.221$ & 3.42 & clean-weak \\
Llama & 8/email & $-1.969$ & $-1.332$ & 1.48 & $+20.642$ & $+17.072$ & 1.21 & clean-weak \\
Llama & 16/name & $-1.060$ & $-0.001$ & 1995 & $+1.465$ & $+0.010$ & 147 & \textbf{clean-strong} \\
Llama & 16/id & $+0.162$ & $-0.046$ & 3.51 & $-0.338$ & $+0.164$ & 2.06 & anti \\
Llama & 16/phone & $+0.043$ & $+0.004$ & 11.02 & $-0.071$ & $+0.007$ & 10.04 & noise \\
Llama & 16/email & $-0.015$ & $+0.005$ & 3.24 & $+0.064$ & $-0.005$ & 12.96 & noise \\
Llama & 24/name & $-0.065$ & $+0.017$ & 3.94 & $+0.127$ & $+0.032$ & 3.98 & noise \\
Llama & 24/id & $+0.189$ & $+0.014$ & 13.42 & $-0.291$ & $-0.015$ & 19.70 & anti \\
Llama & 24/phone & $+0.000$ & $-0.000$ & --- & $+0.000$ & $+0.000$ & --- & inactive \\
Llama & 24/email & $+0.000$ & $+0.000$ & --- & $+0.000$ & $+0.000$ & --- & inactive \\
\hline
Qwen & 9/name & $-1.916$ & $-1.896$ & 1.01 & $+2.530$ & $+2.380$ & 1.06 & clean-weak \\
Qwen & 9/id & $+0.155$ & $-0.196$ & 0.79 & $+3.324$ & $+3.425$ & 0.97 & anti \\
Qwen & 9/phone & $-0.483$ & $-0.770$ & 0.63 & $+11.316$ & $+6.979$ & 1.62 & clean-weak \\
Qwen & 9/email & $+1.106$ & $+0.305$ & 3.62 & $+5.797$ & $+4.995$ & 1.16 & anti \\
Qwen & 18/name & $-1.381$ & $-0.883$ & 1.56 & $+1.972$ & $+1.019$ & 1.94 & clean-weak \\
Qwen & 18/id & $-0.413$ & $-0.068$ & 6.04 & $-0.442$ & $+0.719$ & 0.61 & anti \\
Qwen & 18/phone & $+0.117$ & $+0.327$ & 0.36 & $+0.557$ & $+0.328$ & 1.70 & anti \\
Qwen & 18/email & $-0.442$ & $+0.297$ & 1.49 & $+0.711$ & $+1.076$ & 0.66 & lp-only \\
Qwen & 27/name & $-0.001$ & $-0.000$ & 5.49 & $+0.009$ & $+0.000$ & --- & noise \\
Qwen & 27/id & $-0.014$ & $+0.000$ & 115.1 & $+0.018$ & $-0.005$ & 3.80 & noise \\
Qwen & 27/phone & $-0.169$ & $+0.058$ & 2.92 & $+1.091$ & $-0.013$ & 86.38 & br-only \\
Qwen & 27/email & $-0.093$ & $-0.010$ & 9.51 & $+0.344$ & $+0.064$ & 5.41 & noise \\
\hline
\end{tabular}
\caption{Full per-cell results over 36 configurations. T = target (selected feature); R = coherence-matched random. All values are means over 50 validation pairs $\times$ 9 templates. Inactive configurations have $z_t[f]\approx 0$ across all templates, so neither effect nor selectivity is well-defined.}
\label{tab:full_cell_results}
\end{table*}

\subsection{Per-Template Breakdown}
\label{app:per_template}

This provides the supporting data for the cross-template claims in Section~\ref{sec:results}, including the 9-template consistency of Failure mode A and the OOD-template robustness of Llama L16/name. Due to space constraints, Table~\ref{tab:per_template} reports full per-template values only for cells directly discussed in the main text.

\begin{table}[h]
\centering
\small
\begin{tabular}{lrrr}
\hline
Template & $\Delta_{\mathrm{lp}}^{\mathrm{T}}$ & $\Delta_{\mathrm{lp}}^{\mathrm{R}}$ & $\mathrm{Sel}_{\mathrm{lp}}$ \\
\hline
\multicolumn{4}{l}{\textit{Llama L16/name (clean-strong; positive control)}} \\
T0 (in-distribution) & $-1.153$ & $-0.0003$ & 3791 \\
T1 & $-1.155$ & $-0.0004$ & 3261 \\
T2 & $-1.337$ & $-0.0004$ & 3266 \\
T3 & $-1.346$ & $-0.0004$ & 3549 \\
T4 & $-1.078$ & $-0.0007$ & 1449 \\
T5 (Korean) & $-1.135$ & $+0.0006$ & 1928 \\
T6 & $-1.047$ & $-0.0007$ & 1492 \\
T7 (JSON) & $-0.669$ & $-0.0016$ & 423 \\
T8 (narrative) & $-0.621$ & $-0.0009$ & 692 \\
\hline
\multicolumn{4}{l}{\textit{Gemma L20/name (Failure mode A)}}\\
T0 & $-2.528$ & $-7.548$ & 0.34 \\
T1 & $-2.509$ & $-6.787$ & 0.37 \\
T2 & $-2.415$ & $-7.450$ & 0.32 \\
T3 & $-2.376$ & $-7.378$ & 0.32 \\
T4 & $-3.042$ & $-7.046$ & 0.43 \\
T5 (Korean) & $-3.459$ & $-7.668$ & 0.45 \\
T6 & $-2.760$ & $-7.795$ & 0.35 \\
T7 (JSON) & $-3.152$ & $-7.134$ & 0.44 \\
T8 (narrative) & $-3.002$ & $-8.908$ & 0.34 \\
\hline
\multicolumn{4}{l}{\textit{Llama L8/name (Failure mode B)}} \\
T0 & $-4.552$ & $-1.925$ & 2.37 \\
T1 & $-4.565$ & $-1.939$ & 2.35 \\
T2 & $-4.337$ & $-2.094$ & 2.07 \\
T3 & $-4.592$ & $-1.850$ & 2.48 \\
T4 & $-4.415$ & $-2.730$ & 1.62 \\
T5 (Korean) & $-4.550$ & $-2.075$ & 2.19 \\
T6 & $-4.349$ & $-2.217$ & 1.96 \\
T7 (JSON) & $-5.606$ & $+0.269$ & 20.81 \\
T8 (narrative) & $-4.119$ & $-3.144$ & 1.31 \\
\hline
\end{tabular}
\caption{Per-template results for cells discussed in the main text. T0 retains the original training format containing all four PII fields; T1--T8 isolate a single PII field with varying label formats. Llama L16/name maintains $\mathrm{Sel}_{\mathrm{lp}} > 400$ across all 9 templates including OOD Korean and JSON formats. Gemma L20/name shows $\mathrm{Sel}_{\mathrm{lp}} < 1$ across all 9 templates (Failure mode A).}
\label{tab:per_template}
\end{table}

\subsection{Selected Feature Inventory}
\label{app:selected_feature_inventory}

Table~\ref{tab:selected_features} reports the feature ID selected by the discovery procedure for each (model, layer, PII type). We group the 9 (model, layer) cells by feature-sharing pattern. All full feature-selection collapse cells (Gemma L20, Llama L8, and Qwen L9) fail to pass the clean-strong criterion. The only clean-strong configuration, Llama L16/name, occurs in a non-full-collapse cell. Thus, repeated selection of the same feature across multiple PII types does not guarantee clean-strong feature-specific suppression.

\begin{table}[h]
\centering
\footnotesize
\setlength{\tabcolsep}{3pt}
\begin{tabular}{llcccc}
\hline
Model & Layer & name & id & phone & email \\
\hline
\multicolumn{6}{l}{\textit{Full feature-selection collapse}} \\
Gemma & 20 & 20385 & 20385 & 20385 & 20385 \\
Llama & 8 & 16363 & 16363 & 16363 & 16363 \\
Qwen & 9 & 30639 & 30639 & 30639 & 30639 \\
\hline
\multicolumn{6}{l}{\textit{Partial feature sharing}} \\
Gemma & 9 & 5559 & 5559 & 4648 & 11893 \\
Gemma & 31 & 2807 & 16900 & 2807 & 2807 \\
Llama & 16 & \textbf{32619} & 6432 & 10947 & 10947 \\
Qwen & 18 & 540 & 21146 & 21146 & 21146 \\
\hline
\multicolumn{6}{l}{\textit{No feature sharing}} \\
Llama & 24 & 4823 & 25245 & 5682 & 26877 \\
Qwen & 27 & 27259 & 10595 & 17828 & 5037 \\
\hline
\end{tabular}
\caption{Selected feature ID per (model, layer, PII type). Group definitions: \textit{Full collapse}=single selected feature across all 4 PII types; \textit{Partial sharing}=some PII types share a selected feature; \textit{No sharing}=distinct selected feature for all 4 PII types. SAE dictionary size: $d_{\mathrm{sae}}=28{,}672$ for Gemma, $32{,}768$ for Llama and Qwen. \textbf{Bolded}: feature 32619 (Llama L16/name) corresponds to the only clean-strong configuration in the entire grid.}
\label{tab:selected_features}
\end{table}

As an additional diagnostic, we measure the SAE activation $z_t[f]$ of Llama L8 feature 16363, the clearest full-collapse case, across the four PII types. Its activation magnitude varies by at most 5\% across types (name 6.97, id 7.04, phone 6.77, email 7.01; averaged over 50 validation pairs). Thus, the selected feature appears similar across PII types at both the selection and activation levels, but its intervention effect remains type-biased, as reported under Failure mode B in Section~\ref{sec:results}.

\subsection{Failure Mode A Diagnostic: Gemma L20}
\label{app:pattern_a_diag}

Gemma L20 exhibits Failure mode A, the random-dominant non-selective pattern. For Gemma L20/name, the coherence-matched random direction produces a larger absolute logprob effect than the selected SAE direction in all 50 validation pairs: $|\Delta_{\mathrm{lp}}^{\mathrm{random}}| > |\Delta_{\mathrm{lp}}^{\mathrm{target}}|$ holds in 50/50 pairs. At the per-(pair, template) context level, the same inequality holds in 439/450 contexts (97.6\%). The pattern is observed on both the discovery split ($\mathrm{Sel}_{\mathrm{lp}}=0.25$) and the validation split ($\mathrm{Sel}_{\mathrm{lp}}=0.37$). Because the random control matches the selected-feature intervention in intervention strength, token-wise magnitude, and span-level direction coherence, this random-dominant pattern cannot be attributed to a larger random-vector norm. Rather, it indicates that the selected SAE direction is not more selective than a magnitude-matched random direction in this layer; Appendix~\ref{app:sae_decoder_control} shows that the same holds against decoder directions drawn from the same SAE dictionary. We do not determine whether this arises from the selected feature's failure to generalize, from the dominant PII-related direction not being captured by a single SAE feature, or from the selected direction being weakly misaligned with the intervention-relevant direction.

\subsection{Dictionary-Aligned Control: Random SAE Decoder Directions}
\label{app:sae_decoder_control}

The isotropic random control of Section~\ref{sec:setup} matches the selected-feature intervention in strength, token-wise magnitude, and span-level coherence, but its direction is generated independently of the learned SAE dictionary. In a high-dimensional hidden space, an isotropic unit vector is almost surely nearly orthogonal to the local activation manifold, whereas a decoder direction lies on it by construction. The isotropic control might therefore disrupt next-token prediction more strongly for reasons unrelated to PII, which would make the random-dominant pattern of Failure mode A a geometric artifact rather than evidence about the selected direction.

\paragraph{Control construction.}
To test this, we add a random SAE decoder control. For each selected feature $f$, we sample 20 other non-dead features $g$ from the same model, layer, and SAE dictionary, and replace only the direction while preserving the intervention positions and token-wise magnitude determined by $f$:
\begin{align*}
m_t &= \|z_t[f]\,W_{\mathrm{dec}}[f]\|_2, \\
h'_t &= h_t + \alpha \cdot m_t \cdot \frac{W_{\mathrm{dec}}[g]}{\|W_{\mathrm{dec}}[g]\|_2}.
\end{align*}
The activation $z_t[g]$ of the control feature is not used. The selected intervention and the decoder control are thus matched in position, strength, token-wise magnitude, and span-level coherence, differing only in whether the direction is the selected decoder direction or another direction learned by the same SAE. Non-dead features are those whose \texttt{steps\_since\_active} counter in the SAE checkpoint is at most 1{,}000; the selected feature $f$ is excluded before sampling, and the 20 control features are drawn without replacement with a fixed seed (42). Each control direction is held fixed across the full validation set and evaluated independently, with $\alpha=-2$ as in the main experiments, on all nine prompt templates (T0--T8) and all 50 validation pairs; the reported $\Delta_{\mathrm{lp}}$ for both the selected feature and each control is the mean over these $9 \times 50 = 450$ contexts, matching the aggregation of Table~\ref{tab:full_cell_results}.

\begin{table}[h]
\centering
\small
\setlength{\tabcolsep}{4pt}
\resizebox{\columnwidth}{!}{%
\begin{tabular}{lrrc}
\toprule
Configuration & Selected $\Delta_{\mathrm{lp}}$ & Decoder control $\Delta_{\mathrm{lp}}$ & Stronger \\
\midrule
Llama L16/name & $-1.06$ & $-0.007 \pm 0.014$ & 0/20 \\
Llama L8/name  & $-4.57$ & $-6.60 \pm 3.48$   & 14/20 \\
Qwen L18/name  & $-1.38$ & $-1.04 \pm 0.30$   & 2/20 \\
Gemma L20/name & $-2.80$ & $-9.10 \pm 6.38$   & 20/20 \\
\bottomrule
\end{tabular}%
}
\caption{Random SAE decoder control on four name configurations spanning the main diagnostic categories. Decoder control values are mean $\pm$ s.d.\ over 20 sampled directions. ``Stronger'' is the number of control directions yielding a more negative $\Delta_{\mathrm{lp}}$ (stronger suppression) than the selected direction.}
\label{tab:sae_decoder_control}
\end{table}

\paragraph{Results.}
Table~\ref{tab:sae_decoder_control} reports four name configurations. For Gemma L20/name, the representative Failure mode A case, the random-dominant pattern persists after replacing the isotropic vector with decoder directions from the same SAE: all 20 control directions produce stronger suppression than the selected direction. The Gemma L20 result therefore cannot be explained by a representation-geometry mismatch of the isotropic control alone. For Llama L8/name, 14 of 20 decoder directions produce stronger suppression than the selected direction; since the selected direction had dominated the isotropic control ($\mathrm{Sel}_{\mathrm{lp}}=2.32$, Table~\ref{tab:full_cell_results}), this is an additional finding that the specificity of the selected direction in this feature-selection-collapse cell does not hold up against dictionary-aligned controls. By contrast, for Llama L16/name, the only clean-strong cell, the 20 other decoder directions applied at the same token-wise magnitude show almost no effect, while only the selected direction produces a clear suppression effect ($\Delta_{\mathrm{lp}}=-1.06$); not all SAE interventions are direction-agnostic, and our diagnostic identifies a direction-selective positive case that is distinguishable from matched SAE-direction controls. For Qwen L18/name, the selected direction is stronger than 18 of 20 controls but is not fully separated from their range, so we interpret it as weakly selective rather than random-dominant, consistent with its $\alpha$-dependent selectivity (Appendix~\ref{app:alpha_sensitivity}).

In light of this control, we state the Failure mode A conclusion narrowly: in the random-dominant configurations, the logprob-selected SAE direction was not unusually effective relative to magnitude-matched decoder directions sampled from the same learned SAE dictionary.

\subsection{Failure Mode B Diagnostic: Collapse Cell Asymmetry}
\label{app:collapse_asymmetry}

The three full-collapse cells (Gemma L20, Llama L8, Qwen L9) 
discussed in Section~\ref{sec:results} all exhibit type-asymmetric 
intervention effects despite shared feature selection and matched 
SAE activation magnitudes ($\le 5\%$ variation across PII types 
in each cell). Table~\ref{tab:collapse_asymmetry} summarizes the 
per-type $\Delta_{lp}^{\mathrm{target}}$ values.

\begin{table}[h]
\centering\small
\setlength{\tabcolsep}{4pt}
\begin{tabular}{lcccc}
\toprule
Collapse cell & name & id & phone & email \\
\midrule
Gemma L20 (f=20385) & $-2.80$ & $-0.47$ & $-0.45$ & $-0.31$ \\
Llama L8 (f=16363)  & $-4.57$ & $-1.96$ & $-1.95$ & $-1.97$ \\
Qwen L9 (f=30639)   & $-1.92$ & $+0.16$ & $-0.48$ & $+1.11$ \\
\bottomrule
\end{tabular}
\caption{$\Delta_{lp}^{\mathrm{target}}$ per PII type in the three 
full-collapse cells. Negative values indicate suppression; positive 
values indicate anti-suppression. Despite shared feature selection 
and matched activation magnitudes, intervention effects are markedly 
type-asymmetric: Gemma L20 favors name (6.8$\times$ average of 
others), Llama L8 favors name (2.3$\times$), and Qwen L9 produces 
anti-suppression for id and email.}
\label{tab:collapse_asymmetry}
\end{table}

\subsection{Threshold Sensitivity}
\label{app:threshold_sensitivity}

The ``47\% $\to$ 3\%'' reduction reported in Section~\ref{sec:results} is computed using our default conjunctive criterion: $\Delta_{\mathrm{lp}} \le -0.3$, $\Delta_{\mathrm{br}} \ge +1.0$, $\mathrm{Sel}_{\mathrm{lp}} \ge 5$, $\mathrm{Sel}_{\mathrm{br}} \ge 5$, and no anti-suppression. To test whether this conclusion depends on particular cutoffs, we perform sensitivity analyses along three axes: single-metric selectivity, dual-metric selectivity, and magnitude gate.

\paragraph{Selectivity threshold (single vs dual).}
Table~\ref{tab:threshold_selectivity} reports the number of passing cells as the selectivity threshold is varied, comparing single-metric ($\mathrm{Sel}_{\mathrm{lp}}$ only) and dual-metric ($\mathrm{Sel}_{\mathrm{lp}}$ AND $\mathrm{Sel}_{\mathrm{br}}$) criteria. At $\ge 5$, single-metric admits 2 cells while dual-metric admits 1 (Llama L16/name); the additional cell filtered by dual-metric is Llama L8/phone ($\mathrm{Sel}_{\mathrm{br}} = 3.4$), confirming the necessity of including $\mathrm{Sel}_{\mathrm{br}}$. Even the loosest single-metric threshold ($\ge 2$) admits only 3/36 (8.3\%), far below the 17/36 (47\%) under logprob-only evaluation.

\begin{table}[h]
\centering
\small
\begin{tabular}{lcc}
\hline
Threshold & Single-metric & Dual-metric \\
\hline
$\ge 2$  & 3 (8.3\%) & 3 (8.3\%) \\
$\ge 3$  & 2 (5.6\%) & 2 (5.6\%) \\
$\ge 5$  & 2 (5.6\%) & 1 (2.8\%) \\
$\ge 10$ & 2 (5.6\%) & 1 (2.8\%) \\
\hline
\end{tabular}
\caption{Selectivity threshold sensitivity. Single-metric: $\mathrm{Sel}_{\mathrm{lp}}$ only; Dual-metric: both $\mathrm{Sel}_{\mathrm{lp}}$ and $\mathrm{Sel}_{\mathrm{br}}$ above threshold. Default magnitude gates ($\Delta_{\mathrm{lp}} \le -0.3$, $\Delta_{\mathrm{br}} \ge +1.0$) and anti-suppression filter retained. Single-metric at $\ge 5$ retains Llama L8/phone ($\mathrm{Sel}_{\mathrm{lp}} = 11.2$, $\mathrm{Sel}_{\mathrm{br}} = 3.4$), which dual-metric filters out.}
\label{tab:threshold_selectivity}
\end{table}

\paragraph{Magnitude gate.}
Table~\ref{tab:threshold_magnitude} reports the number of passing cells when we vary the magnitude gate while keeping the selectivity threshold fixed at $\ge 5$.

\begin{table}[h]
\centering
\small
\begin{tabular}{lcc}
\hline
$|\Delta_{\mathrm{lp}}^{\mathrm{T}}|$ gate & $\Delta_{\mathrm{br}}^{\mathrm{T}}$ gate & Cells / 36 \\
\hline
$\ge 0.0$ & $\ge 0.0$ & 2 \\
$\ge 0.0$ & $\ge 0.5$ & 1 \\
$\ge 0.3$ & $\ge 1.0$ (default) & 1 \\
$\ge 0.5$ & $\ge 1.0$ & 1 \\
\hline
\end{tabular}
\caption{Magnitude gate sensitivity at fixed $\mathrm{Sel} \ge 5$. Even the loosest gates yield only 2/36 (5.6\%) clean-strong cells.}
\label{tab:threshold_magnitude}
\end{table}

Across all three sensitivity axes, even the loosest setting admits at most 3/36 clean-strong cells (8.3\%). Thus, the central conclusion of this work is robust to threshold choice: the 47\% apparent success rate under logprob-only evaluation collapses to a single clean-strong cell under strict multi-metric evaluation.

\subsection{$\alpha$ Sensitivity}
\label{app:alpha_sensitivity}

We fix $\alpha=-2$ in all main experiments. To verify that our 
diagnostic conclusions are not artifacts of this choice, we sweep 
$\alpha \in \{-1, -2, -3, -4, -5\}$ on three cells representing 
distinct categories: Llama L16/name (clean-strong), Gemma L20/name 
(Failure mode A), and Qwen L18/name (clean-weak). All other settings 
(model, layer, feature, validation pairs) are fixed; templates are 
restricted to T1 and T5 (one in-distribution, one Korean OOD). 
The $\alpha=-2$ values match the corresponding 9-template means 
in Table~\ref{tab:full_cell_results} exactly.

\begin{figure*}[t]
\centering
\includegraphics[width=\linewidth]{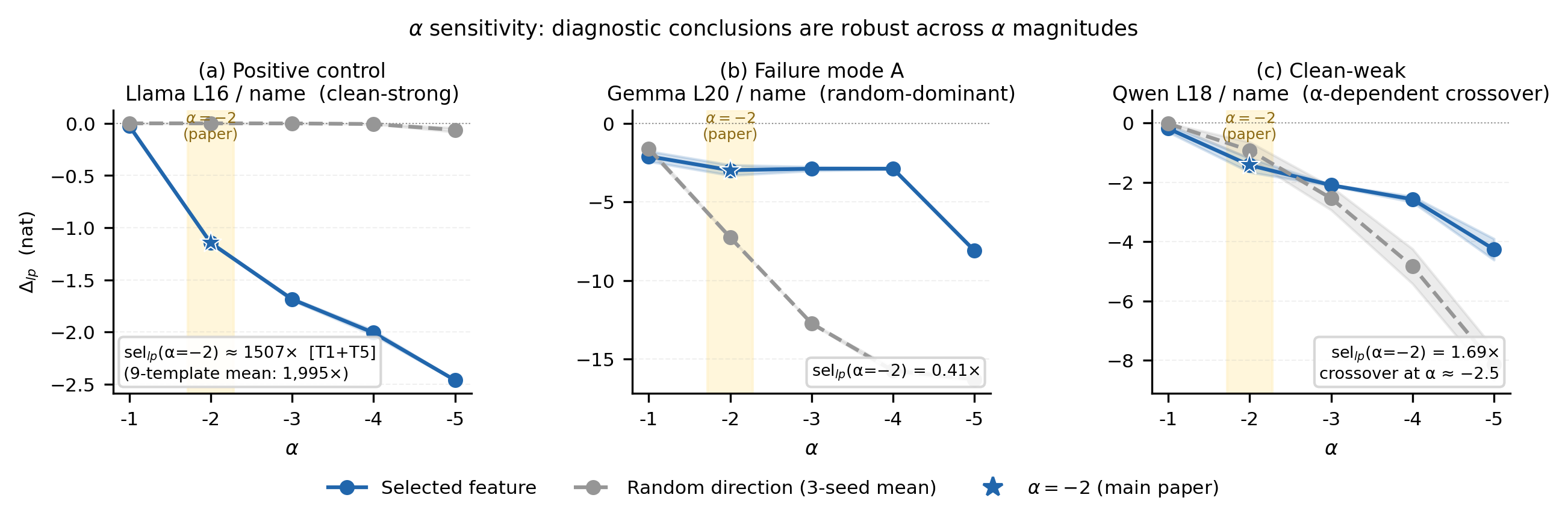}
\caption{$\alpha$ sensitivity on three representative cells. 
Yellow band: main-paper setting ($\alpha=-2$). 
(a) Llama L16/name (clean-strong): selected-feature dominance is 
preserved across all tested $\alpha$. 
(b) Gemma L20/name (Failure mode A): random-direction dominance is 
preserved across all $\alpha$, including the over-perturbation 
regime at $\alpha=-5$. 
(c) Qwen L18/name (clean-weak): $\mathrm{sel}_{lp}$ exceeds 1 at 
$\alpha=-1, -2$ but crosses below 1 from $\alpha=-3$ onward, 
indicating that the weak selectivity observed in the main experiment 
is itself $\alpha$-dependent.}
\label{fig:alpha_sensitivity}
\end{figure*}

Figure~\ref{fig:alpha_sensitivity} shows that both the clean-strong 
classification (Llama L16/name) and Failure mode A (Gemma L20/name) 
are $\alpha$-robust: selected-feature dominance and random-direction 
dominance, respectively, hold across all tested $\alpha$, including 
the over-perturbation regime at $\alpha=-5$.

Table~\ref{tab:alpha_sensitivity} additionally examines Qwen L18/name as 
a transitional case: $\mathrm{sel}_{lp}$ exceeds 1 at $\alpha=-1, -2$ 
but flips below 1 from $\alpha=-3$ onward. This indicates that the 
weak selectivity observed in the main experiment is itself 
$\alpha$-dependent, and that $\alpha=-2$ if anything overestimates 
rather than underestimates selectivity for such cells.

\begin{table}[t]
\centering
\small
\setlength{\tabcolsep}{5pt}
\begin{tabular}{lccccc}
\toprule
\multicolumn{6}{c}{$\mathrm{sel}_{lp}$ as a function of $\alpha$ (T1+T5)} \\
\midrule
Cell & $-1$ & $-2$ & $-3$ & $-4$ & $-5$ \\
\midrule
Llama L16/name  & ---   & 1507 & 2486 & 320  & 39   \\
Gemma L20/name  & 1.31  & 0.41 & 0.23 & 0.18 & 0.50 \\
Qwen L18/name   & 1.31  & 1.54 & \textbf{0.83} & 0.53 & 0.52 \\
\bottomrule
\end{tabular}
\caption{Selectivity ratio $\mathrm{sel}_{lp}$ across $\alpha$ for 
three representative cells: Llama L16/name (clean-strong), Gemma 
L20/name (Failure A), Qwen L18/name (clean-weak). Bold marks 
the $\alpha$ at which selectivity flips below 1. Llama L16/name 
$\alpha=-1$ omitted ($|\Delta_{lp}^\text{random}|$ at measurement floor).}
\label{tab:alpha_sensitivity}
\end{table}

\subsection{Downstream Utility}
\label{app:utility}

To test whether the configurations separated by our multi-metric diagnostic also diverge in downstream behavior, we evaluate three configurations representing the main diagnostic categories, Llama L16/name (clean-strong), Qwen L18/name (clean-weak), and Gemma L20/name (Failure mode A), on PubMedQA, MedQA, and MMLU before and after intervention. Identical evaluation examples and protocols are used in all conditions, with the intervention applied exactly as in the main experiments ($\alpha=-2$, selected feature fixed). The three configurations were chosen as the representative cells of the clean-strong, clean-weak, and Failure mode A categories that are also used in the $\alpha$ sensitivity analysis (Appendix~\ref{app:alpha_sensitivity}).

\paragraph{Protocol.}
Each benchmark is evaluated on 200 examples: for PubMedQA, the first 200 examples of the expert-labeled (PQA-L) test split; for MedQA (English 4-option USMLE test set), 200 examples sampled uniformly at random with seed 42; for MMLU, 200 examples sampled uniformly at random with seed 42 from the full test set across all subjects. All prompts are zero-shot. For the two instruction-tuned models (Llama-3.1-8B-Instruct and Qwen3-8B), we use the chat template with a system instruction that requests exactly one answer token (\texttt{yes}/\texttt{no}/\texttt{maybe} for PubMedQA; \texttt{A}--\texttt{D} otherwise), with Qwen3's thinking mode disabled; for Gemma-2-9B we use a plain completion prompt ending in ``\texttt{Answer:}''. PubMedQA contexts are truncated to the first two abstracts of at most 600 characters each. No text is generated: the prediction is the argmax over the logits of the admissible label tokens at the final position of a single forward pass, so no answer parsing is required and every example receives a prediction. The intervention is applied as a forward hook on the full output of the selected layer; because the intervention term is proportional to $z_t[f]$, it vanishes automatically at positions where the selected feature is inactive, so no PII-span mask is needed for benchmark prompts. The same examples and per-example seeds are used for the baseline and intervention conditions.

\begin{table}[h]
\centering
\small
\setlength{\tabcolsep}{4pt}
\resizebox{\columnwidth}{!}{%
\begin{tabular}{lccc}
\toprule
Configuration & PubMedQA & MedQA & MMLU \\
\midrule
Llama L16/name (clean-strong) & $31.0 \to 32.0$ & $45.5 \to 46.5$ & $61.0 \to 61.5$ \\
Qwen L18/name (clean-weak)    & $28.5 \to 16.5$ & $57.0 \to 23.5$ & $71.5 \to 48.5$ \\
Gemma L20/name (Failure A)    & $47.5 \to 33.5$ & $60.5 \to 34.5$ & $67.5 \to 45.0$ \\
\bottomrule
\end{tabular}%
}
\caption{Downstream accuracy (\%) before $\to$ after intervention. Llama L16/name, the only clean-strong cell, shows no degradation on any benchmark, whereas both non-clean-strong cells degrade consistently across all three.}
\label{tab:utility}
\end{table}

Llama L16/name shows no performance degradation relative to the no-intervention baseline on any benchmark (Table~\ref{tab:utility}). The two non-clean-strong cases degrade consistently across all three benchmarks, by 12 to 34 accuracy points. This supports the point that PII logprob reduction does not by itself indicate selective, utility-preserving suppression, and that the cells distinguished by our multi-metric diagnostic also diverge in actual downstream behavior. It also bears on the interpretation of extraction-based evaluation: an intervention that lowers extraction success while degrading general generation, as in Qwen L18/name and Gemma L20/name, cannot be distinguished from a PII-specific one by extraction rate alone, which is why we frame extraction benchmarks as complementary to, rather than a substitute for, feature-specificity diagnostics.

\subsection{Causal Mediation Feasibility Study}
\label{app:mediation_study}

This appendix provides the full protocol and results of the feasibility study summarized in Section~\ref{sec:research_mission}. Rather than merely proposing the causal mediation analysis framework of \citet{vig2020investigating} as future work, the feasibility study demonstrates that it is actually implementable in the PII suppression setting, and examines whether correlation-selected SAE features pass causal verification.

\paragraph{Pair construction.}
For each model, layer, and PII type, we construct 50 pairs of text containing the original PII ($x_1$) and counterfactual text ($x_0$) in which the PII is replaced with a different value of the same type. We write $\mathrm{lp}(y \mid x)$ for the log-probability of the original PII span $y$ given text $x$, so that $\mathrm{lp}(y \mid x_1) - \mathrm{lp}(y \mid x_0)$ measures how much the model prefers the original PII value over the counterfactual.

\paragraph{Intervention position and noising-style intervention.}
The intervention position is the token immediately preceding the PII span. Because the preceding text in $x_0$ and $x_1$ is identical, the activations themselves are the same at that position, and applying the original NIE definition of \citet{vig2020investigating} as-is (patching in the activation of $x_1$ during a run on $x_0$) causes the metric to become identically zero. To avoid this, we instead use a noising-style intervention that zero-ablates the relevant feature during the $x_1$ run, which conceptually corresponds to the noising intervention used in the causal tracing of \citet{meng2022locating}. The control condition ($\mathrm{NIE}_{\mathrm{rand}}$) uses a random direction of matched magnitude.

\paragraph{Effect definitions.}
With $\mathrm{lp}(y \mid x_1, f{=}0)$ denoting the log-probability of the original PII span under the zero-ablation of $f$, we define
\begin{align*}
\mathrm{TE} &= \mathrm{lp}(y \mid x_1) - \mathrm{lp}(y \mid x_0), \\
\mathrm{NIE}_f &= \mathrm{lp}(y \mid x_1, f{=}0) - \mathrm{lp}(y \mid x_0), \\
\mathrm{MF}_f &= 1 - \mathrm{mean}(\mathrm{NIE}_f) / \mathrm{mean}(\mathrm{TE}),
\end{align*}
where the means are taken over the 50 pairs of a cell, so that the mediation fraction $\mathrm{MF}_f$ is computed as a ratio of effects averaged across pairs. $\mathrm{NIE}_{\mathrm{rand}}$ is defined analogously for the random direction. Note that $\mathrm{NIE}_f$ as used here denotes the effect that remains after ablating $f$, and the share of the total effect mediated by $f$ is given by $\mathrm{MF}_f$. TE is defined independently of any feature intervention or aggregation choice, and serves as a memorization signal. A large positive TE indicates that the model assigns substantially higher probability to the original PII value than to a counterfactual one.

\paragraph{Candidate selection.}
Unlike the logprob-driven selection used in the main experiments, candidate features here are selected independently of logprob: for each (model, layer, PII type), we take the five non-dead SAE features with the highest activation rate at the position immediately preceding the PII span across the $x_1$ texts. This design decouples feature selection from evaluation, so that logprob is not used for both simultaneously. The target models are the same three models as in the main experiments (Gemma-2-9B, Llama-3.1-8B-Instruct, Qwen3-8B), and we run 3 models $\times$ 3 layers $\times$ 4 PII types $\times$ 5 candidate features $=$ 180 cells with 50 pairs per cell, for a total of 9,000 causal patching runs.

\paragraph{Memorization by PII type.}
Table~\ref{tab:study_te} reports TE for all 12 (model, PII type) combinations. Email shows a clear memorization signal ($\mathrm{TE} > 2.4$) across all three models, whereas phone number and patient ID show largely no memorization, and name shows inconsistent results depending on the model.

\begin{table}[h]
\centering
\small
\begin{tabular}{lrrr}
\toprule
PII type & Gemma-2-9B & Llama-3.1-8B & Qwen3-8B \\
\midrule
name  & $+0.609$ & $-0.283$ & $-0.604$ \\
id    & $-0.413$ & $-0.233$ & $-0.025$ \\
phone & $-0.038$ & $+0.067$ & $-0.065$ \\
email & $+2.580$ & $+3.560$ & $+2.466$ \\
\bottomrule
\end{tabular}
\caption{Total effect (TE) by PII type and model, averaged over 50 counterfactual pairs per (model, PII type) cell. TE is defined independently of any feature intervention or aggregation choice.}
\label{tab:study_te}
\end{table}

\paragraph{Indistinguishability from the random control.}
Table~\ref{tab:mediation_study} reports, for each (model, PII type) combination with $|\mathrm{TE}| \ge 0.05$, the mean $\mathrm{NIE}_f$ and $\mathrm{NIE}_{\mathrm{rand}}$ averaged across the 15 candidate features (3 layers $\times$ 5 features), their ratio, and the fraction of the 15 features whose individual $\mathrm{NIE}_f / \mathrm{NIE}_{\mathrm{rand}}$ falls in $(0.5, 2.0)$, which we call indistinguishable. The core finding of the main selectivity analysis, that most selected features are not superior to random directions, is corroborated under this fully independent causal mediation methodology: of 150 candidate features with significant TE, 146 (97.3\%) are indistinguishable from random. When broken down by (model, PII type), all candidates are indistinguishable from random in 9 of 10 combinations, with the sole exception being Gemma-2-9B/name (11/15, 73.3\%), precisely the one combination exhibiting a positive memorization signal ($\mathrm{TE} > 0$). This indicates that causal mediation analysis performs correctly, reliably distinguishing candidates according to whether genuine memorization is present. The limitation therefore lies not in verification but in candidate generation.

\paragraph{Disagreement between feature selection criteria.}
Comparing the logprob-based selection of the main experiments against the activation-rate-based selection of the feasibility study across all 36 (model, layer, PII type) combinations, we find zero overlapping features. The two selection criteria therefore measure entirely different axes of the feature space, sensitivity of output change versus input-selective activation, and logprob-based selection is not only circular but may also miss the actual causal bottleneck. Furthermore, the fact that candidate pools drawn from these completely different criteria still show a similarly low rate of passing causal verification lends further support to the argument in Section~\ref{sec:research_mission} that the causal pathways underlying PII leakage need to be investigated directly, by combining circuit-level localization with causal mediation analysis.

\end{document}